\documentclass[letterpaper]{article} 
\usepackage{aaai25}  
\usepackage{times}  
\usepackage{helvet}  
\usepackage{courier}  
\usepackage[hyphens]{url}  
\usepackage{graphicx} 
\usepackage{natbib}  
\usepackage{caption} 
\usepackage{algorithm}
\usepackage{algorithmic}

\usepackage{booktabs}
\usepackage{multirow}
\usepackage{amsmath}
\usepackage{subfigure}
\usepackage{xcolor}

\usepackage{newfloat}
\usepackage{listings}
\DeclareCaptionStyle{ruled}{labelfont=normalfont,labelsep=colon,strut=off} 
\floatstyle{ruled}
\newfloat{listing}{tb}{lst}{}
\floatname{listing}{Listing}
\title{DLF: Disentangled-Language-Focused Multimodal Sentiment Analysis}
\author{
    Pan Wang\textsuperscript{\rm 1},
    Qiang Zhou\textsuperscript{\rm 1},
    Yawen Wu\textsuperscript{\rm 1},
    Tianlong Chen\textsuperscript{\rm 2},
    Jingtong Hu\textsuperscript{\rm 1}
}
\affiliations{
    \textsuperscript{\rm 1}University of Pittsburgh\\
    \textsuperscript{\rm 2}UNC Chapel Hill\\
    \{pan.wang, qiang.zhou, yawen.wu, jthu\}@pitt.edu,
    tianlong@cs.unc.edu
    
}

\usepackage{bibentry}

\begin{document}

\maketitle

\begin{abstract}
Multimodal Sentiment Analysis (MSA) leverages heterogeneous modalities, such as language, vision, and audio, to enhance the understanding of human sentiment. While existing models often focus on extracting shared information across modalities or directly fusing heterogeneous modalities, such approaches can introduce redundancy and conflicts due to equal treatment of all modalities and the mutual transfer of information between modality pairs. To address these issues, we propose a Disentangled-Language-Focused (DLF) multimodal representation learning framework, which incorporates a feature disentanglement module to separate modality-shared and modality-specific information. To further reduce redundancy and enhance language-targeted features, four geometric measures are introduced to refine the disentanglement process. A Language-Focused Attractor (LFA) is further developed to strengthen language representation by leveraging complementary modality-specific information through a language-guided cross-attention mechanism. The framework also employs hierarchical predictions to improve overall accuracy. Extensive experiments on two popular MSA datasets, CMU-MOSI and CMU-MOSEI, demonstrate the significant performance gains achieved by the proposed DLF framework. Comprehensive ablation studies further validate the effectiveness of the feature disentanglement module, language-focused attractor, and hierarchical predictions. 
\end{abstract}

\begin{links}
    \link{Code}{https://github.com/pwang322/DLF.}
\end{links}

\section{Introduction}
\label{sec:intro}
\begin{figure}[t]
  \centering
  \includegraphics[width=1\linewidth]{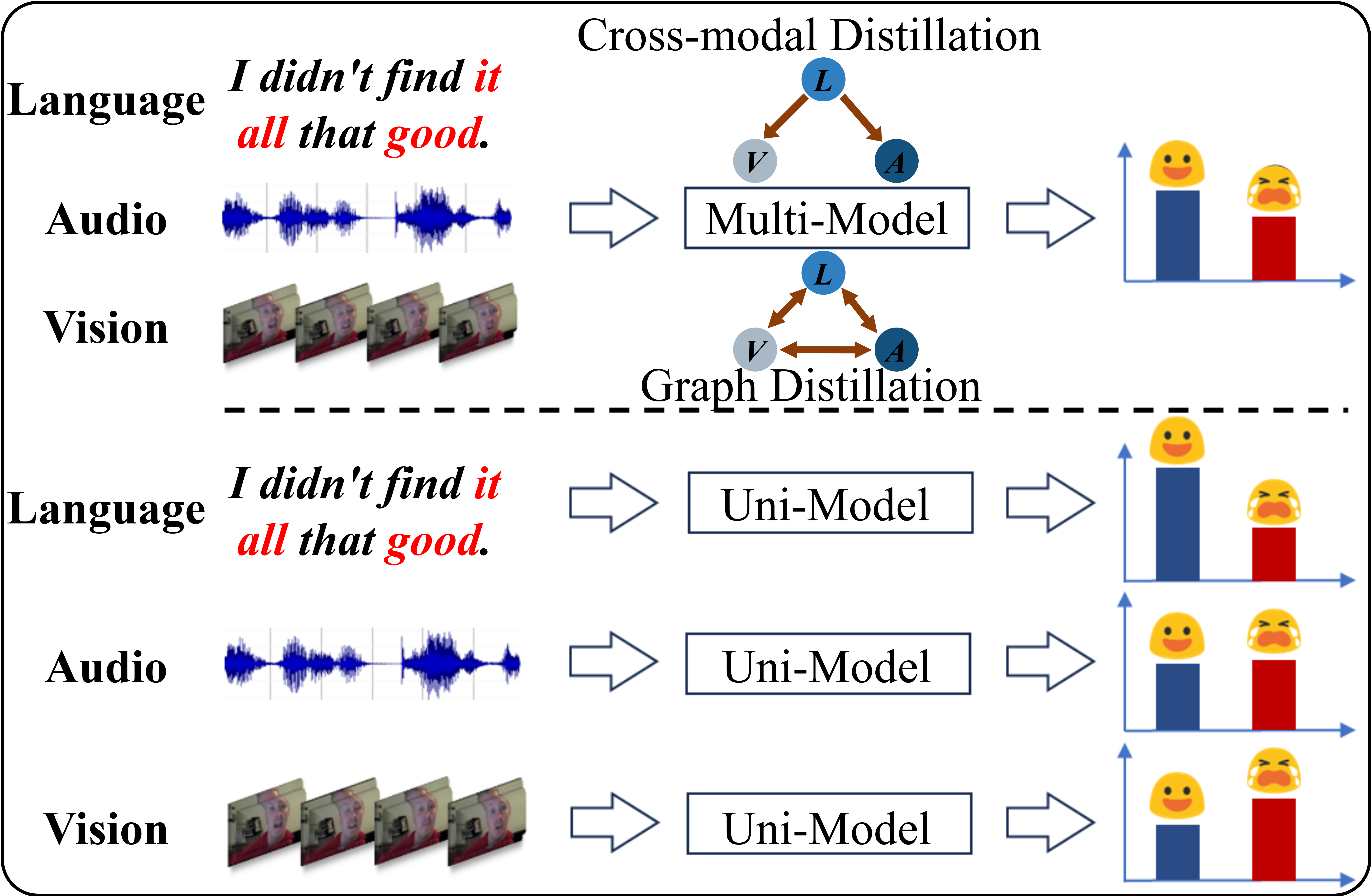}
  \caption{Task pipeline of the Multimodal Sentiment Analysis, 
  and varied performance of different modalities.}
  \label{fig:bg}
\end{figure}
With the rapid development of social media, multimodal interaction has become increasingly popular, which attracts many researchers to transfer uni-modal learning to multi-modal learning tasks \cite{awal2024vismin,guan2024hallusionbench,xu2023multimodal}. One of the most significant subfields is multimodal sentiment analysis (MSA) \cite{geetha2024multimodal,yang2022disentangled}.
MSA aims to perceive human sentiment through multiple heterogeneous modalities, such as language, vision, and audio, playing a crucial role in many applications including cognitive psychology, scenario understanding, and mental health  \cite{ali2023unified, ezzameli2023emotion,yang2023context}.  Compared with unimodal solutions, MSA often presents a more robust performance by leveraging complementary information from different modalities. How to effectively learn essential representations without redundant and conflicting information from multiple heterogeneous modalities, however, remains an open question in the academic field, especially in multimodal learning communities.

In recent years, researchers have shown an increased interest in MSA. Many multimodal models have been proposed to facilitate MSA, and they can be categorized into two groups:
representation learning-oriented methods \cite{guo2022dynamically,hazarika2020misa,sun2023efficient,yang2022learning} and multimodal fusion-oriented methods \cite{zhang2023learning,huang2020multimodal,yang2022contextual,tsai2019multimodal,lv2021progressive,rahman2020integrating}. 
The former primarily aims to acquire an advanced semantic understanding of various modalities enriched with diverse clues of human sentiments, resulting in more powerful human sentiment encoders. Conversely, the latter emphasizes designing sophisticated fusion strategies at various levels, including feature-level, decision-level, and model-level fusion, to derive unified representations from multimodal data. It is worth noting that the fundamental aspect of MSA lies in learning and integrating multimodal representations, where the goal is to accurately process and integrate various modal inputs to discern sentiments from the underlying data. Although current leading methods in MSA \cite{hazarika2020misa, tsai2019multimodal,zadeh2017tensor, yu2021learning} have shown considerable progress, the inherent disparities across diverse modalities continue to present challenges, complicating the development of stable and effective multimodal representation.
For MSA, as shown in Figure \ref{fig:bg}, existing works and our ablation study (see Table \ref{tab:A2}) have shown that language, vision, and audio sources contribute differently to the overall prediction performance \cite{pham2019found, tsai2019multimodal, kim2023aobert, li2024unified,lei2023text}, which indicates the big distribution gap among different modalities hinders the final performance.

To mitigate the distribution gap among heterogeneous modalities,  as shown in Figure \ref{fig:bg}, knowledge distillation-based methods, such as cross-modal and graph distillation, are introduced to transfer reliable information between different modalities \cite{gupta2016cross, guo2020online,aslam2023privileged,kim2022cross, hazarika2020misa,li2023decoupled, tsai2019multimodal}. Cross-modal distillation typically leverages the stronger \textit{Language} modality to teach weaker modalities (\textit{Vision and Audio}), while graph distillation completely performs bidirectional information transfer between all modality pairs. However, it is important to note that conventional distillation is inherently asymmetric—it is effective when transferring information from one modality to another, but the benefits of the reciprocal process are unclear. This asymmetry can lead to redundant or even conflicting information in cross-modal and graph distillation, ultimately limiting overall performance.

To this end, we critically reconsider the characteristics illustrated in Figure \ref{fig:bg}: \textit{\textbf{Why focus solely on bridging the gap between different modalities, rather than strategically enhancing the strengths of the dominant one?}} However, directly enhancing the dominant modality while treating all modalities equally and employing bidirectional information transfer across all modality pairs often introduces redundancy and conflicts \cite{hazarika2020misa}, thereby reducing overall performance.
In contrast, our work strategically leverages a pivotal characteristic of MSA: language has been empirically recognized as the dominant modality \cite{tsai2019multimodal}. Building on this insight, we intend to develop a novel Language-Focused Attractor (LFA), a targeted enhancement scheme designed to transfer complementary information exclusively to the dominant language modality, which consolidates information through pathways such as Video $\rightarrow$ Language, Audio $\rightarrow$ Language, and Language $\rightarrow$ Language, resulting in effectively minimizing redundancy and conflicting information and improving overall MSA accuracy.

To achieve this, we propose a Disentangled-Language-Focused (DLF) multimodal representation learning framework to fully exploit the potential of language-dominant MSA. The framework follows a structured pipeline: feature extraction, disentanglement, enhancement, fusion, and prediction.
To specifically address the issues of redundancy and conflicting information to facilitate language-targeted feature enhancement, DLF introduces four geometric measures as regularization terms in the total loss function, effectively refining shared and specific spaces both separately and jointly. Within the modality-specific space, we further develop the LFA to enhance language representation by attracting complementary information from other modalities. This process is guided by a \textit{Language-Query}-based multimodal cross-attention mechanism, ensuring precise and targeted feature enhancement between heterogeneous modality pairs ($X$$\rightarrow$Language, where $X$ refers to Language, Video, or Audio).
Finally, the enhanced shared and specific features are fused, followed by hierarchical predictions to further improve overall prediction accuracy.

Our main contributions can be summarized as follows:
\begin{itemize}
    \item \textbf{Proposed Framework:} In this study, we propose a Disentangled-Language-Focused (DLF) multimodal representation learning framework to promote MSA tasks. The DLF framework presents a structured pipeline: feature extraction, disentanglement, enhancement, fusion, and prediction.
    \item \textbf{Language-Focused Attractor (LFA):} We develop the LFA to fully harness the potential of the dominant language modality within the modality-specific space. The LFA exploits the language-guided multimodal cross-attention mechanisms to achieve a targeted feature enhancement ($X$$\rightarrow$Language).
    \item \textbf{Hierarchical Predictions:} We devise hierarchical predictions to leverage the pre-fused and post-fused features, improving the total MSA accuracy. Comprehensive ablation studies further validate the effectiveness of each component in the DLF framework.
\end{itemize}

\begin{figure*}[t]
  \centering
  \includegraphics[width=1.0\linewidth]{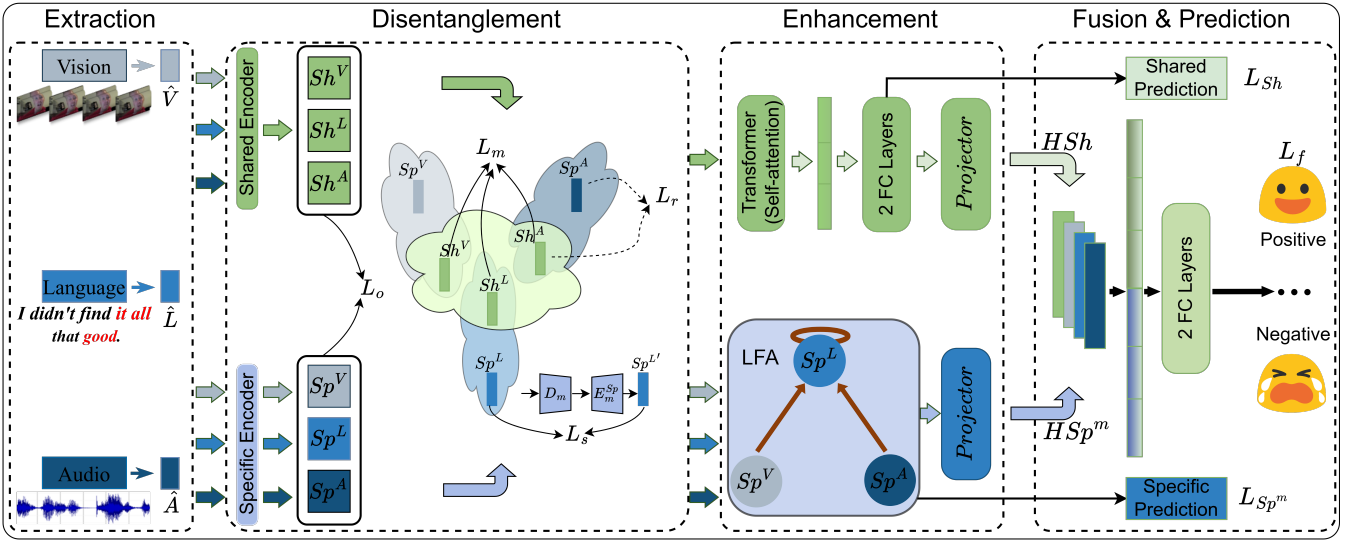}
  \caption{Overview of the proposed DLF framework. The framework follows a pipeline of feature extraction, disentanglement, enhancement, fusion, and prediction, featuring three core components: the feature disentanglement module, the Language-Focused Attractor (LFA), and hierarchical predictions (including shared prediction, specific prediction,
and final prediction).}
  \label{fig:Framework}
\end{figure*}
\section{Related Work}
\label{sec:related works}
\subsection{Multimodal Sentiment Analysis}
Multimodal Sentiment Analysis (MSA) integrates information from diverse modalities, such as language, video, and audio \cite{ali2023unified, ezzameli2023emotion}. Mainstream methods can be categorized into representation learning-oriented \cite{guo2022dynamically,sun2023efficient,yang2022learning} and fusion-oriented approaches \cite{zhang2023learning,huang2020multimodal,tsai2019multimodal}. Representation methods like \cite{guo2022dynamically}, enhance cross-modal interactions, while fusion techniques, including Transformer-based model \cite{huang2020multimodal}, focus on combining features effectively. Despite progress, performance disparities among modalities hinder the overall prediction accuracy.
To this end, distillation-based strategies were introduced to bridge the gap. For instance,  \citet{kim2022cross} proposed cross-modal distillation between textual and auditory modalities to enhance emotion classification granularity. To dynamically adapt to distillation, ternary-symmetric architectures (MulT) \cite{tsai2019multimodal} or graph distillation units \cite{zadeh2018multimodal,li2023decoupled} were introduced, representing modalities as vertices and their interactions as edges. Recent approaches also leverage large multimodal language models for flexible interactions \cite{wu2023next} and integrate contextual knowledge to boost predictions \cite{wang2024wisdom}. However, previous paradigms treat different modalities equally, easily causing redundant and conflicting information. Our proposed Language-Focused Attractor (LFA) directs knowledge transfer to the dominant language, mitigating redundant and conflicting information and enhancing overall performance.

\subsection{Disentangled Multimodal Representation Learning}
Disentangled multimodal representation learning aims to separate distinct factors of variation across multiple modalities (e.g., vision, language, audio) into independent subspaces \citep{yang2023what2comm,li2024correlation,yang2023context}. \citet{tsai2018learning} introduced the concept of factorized representations for multimodal data, designed to disentangle modality-specific and shared components, successfully isolating audio and visual features in audiovisual speech recognition. Building on this, \citet{hazarika2020misa} proposed the MISA framework, which projects each modality into common and private feature spaces, reducing inter-modality disparities while enhancing representation diversity. Further advancements including \citet{yang2022disentangled,yang2022learning}, employed metric learning and adversarial learning to construct modality-invariant and modality-specific subspaces, significantly improving multimodal fusion. \citet{li2023decoupled} developed a decoupled multimodal distillation (DMD) approach to address distribution gaps between modalities. However, previous methods uniformly treat all modalities regarding the purpose of disentanglement. In our DLF, disentanglement aims to facilitate language-targeted feature enhancement in LFA, thus, we introduce four measures as regularization terms, carefully designed to reinforce disentanglement by jointly and independently optimizing shared and specific spaces.

\section{Proposed Approach}
\label{sec:method}

\textbf{Preliminaries.}
As shown in Figure \ref{fig:Framework}, the task of MSA aims to predict the sentiment intensity or label of given multimodal inputs. In DLF, three modalities are concurrently considered, such as Language ($L$), Vision ($V$), and Audio ($A$), represented as 2D tensors $\hat{X}_m\in{R}^{N_m\times d_m}$, where $N_m$ is the sequence length, $d_m$ is the embedding dimension, and $m\in\{L, V, A\}$ means different modalities.

\subsection{Overview}
\label{sec:overview}
The framework of the proposed DLF is illustrated in Figure \ref{fig:Framework}. It adopts a structured pipeline comprising feature extraction, disentanglement, enhancement, fusion, and prediction. The framework integrates three core components: the feature disentanglement module, the Language-Focused Attractor (LFA), and hierarchical predictions (shared, specific, and final predictions).
DLF decomposes multimodal features into modality-shared and modality-specific spaces to minimize redundancy and conflicts among heterogeneous modalities. To reinforce this decoupling, four geometric measures are incorporated into the total loss as regularization terms. Additionally, the LFA is designed to leverage the dominant language modality by integrating complementary information from other modalities, thereby enhancing language representation. Finally, hierarchical predictions are performed to boost overall MSA performance. The details are as follows:

\subsection{Feature Disentanglement Module}
\label{sec:Decoupling}
To reduce redundant and conflicting information, the proposed DLF framework utilizes a shared encoder and three modality-specific encoders to decompose multimodal information into modality-shared and modality-specific feature spaces, denoted as $Sh^m$ and $Sp^m$, respectively, where $m \in \{V, L, A\}$. Formally, the shared and specific encoders are defined as:

\begin{equation} 
Sh^m = E_m^{Sh}(\hat{X}_m), 
\label{eq:Sh}
\end{equation} 
\begin{equation} 
Sp^m = E_m^{Sp}(\hat{X}_m), 
\label{eq:Sp} 
\end{equation}
where $E_m^{Sh}$ and $E_m^{Sp}$ represent the shared and specific encoders, respectively. In this work, both encoders are implemented as cascaded Transformer layers.

For effective disentanglement, DLF incorporates the regularization effect of carefully designed regularization terms. While classical approaches often employ distribution similarity measures such as KL-Divergence along the hidden dimensions \cite{kim2018disentangling}, we adopt four geometric measures based on Euclidean distances and cosine similarity due to their intuitive nature and computational efficiency.

After initial disentanglement, DLF concatenates $Sh^m$ and $Sp^m$ for each modality and reconstruct the multimodal input $\hat{X}_m$, resulting $\hat{X}_m^{\prime}$ by decoding the fused features [$Sh^m$$\oplus$$Sp^m$].
This process can be formulated as follows:
\begin{equation}
  \hat{X}_m^{\prime} = D_m([Sh^m \oplus Sp^m]),
  \label{eq:X'}
\end{equation}
where $D_m$ is the 1D convolution decoder, and $\oplus$ is the concatenation operation.
The discrepancy between the low-level features $\hat{X}_m$ and the generated features $\hat{X}_m^{\prime}$, called the reconstruction loss $L_r$, can serve as a regularization term contributing to the feature disentanglement module:
\begin{equation}
  L_r = ||\hat{X}_m - \hat{X}_m^{\prime}||^2.
  \label{eq:Lr}
\end{equation}
Furthermore, the modality-specific reconstruction process can be formulated as:
\begin{equation}
  {Sp^m}^{\prime} = E_m^{Sp}(\hat{X}_m^{\prime}),
  \label{eq:Spm'}
\end{equation}
where ${Sp^m}^{\prime}$ is estimated modality-specific features from the modality-specific reconstruction process. Naturally, the discrepancy between original modality-specific features $Sp^m$ and the estimated ones ${Sp^m}^{\prime}$ can be regarded as the specific loss $L_s$:
\begin{equation}
  L_s = || Sp^m-{Sp^m}^{\prime}||^2.
  \label{eq:Ls}
\end{equation}

Although the reconstruction loss $L_r$ and specific loss $L_s$ contribute to the decoupling process, their effectiveness in achieving robust disentanglement remains limited. This limitation arises from the potential sub-optimal performance of the shared encoder during the initial training phase, especially when compared to the modality-specific encoder. Such a disparity, if left unaddressed, is exacerbated by these loss functions, causing an increasing divergence between the two encoders as training progresses.
Therefore, we incorporate a modified triplet loss \cite{schroff2015facenet} to enhance the performance of the modality-shared encoder. The triplet loss is defined as:
\begin{equation} 
\begin{aligned} 
L_m = \frac{1}{|T|} \max\left(0, d\left(S, P\right) - d\left(S, N\right) + \mu\right), \end{aligned} 
\label{eq:Lm} 
\end{equation}
where $S$ represents a sampled modality in the modality-shared space, $P$ denotes the positive sample corresponding to the representation of the same sentiment across different modalities, and $N$ refers to the negative sample representing distinct sentiments within the same modality. $T$ is the total number of positive and negative samples, $d(\cdot, \cdot)$ computes the cosine similarity between two feature vectors, and $\mu$ represents a distance margin.

The aforementioned losses, $L_r$, $L_s$, and $L_m$, regulate the shared and specific features to ensure they focus on their respective objectives. To further refine the decoupling between these two spaces, a soft orthogonality loss, $L_o$, is introduced to minimize redundancy and conflicts between shared and specific multimodal features. It is defined as:

\begin{equation} 
L_o = O(Sh^m, Sp^m), 
\label{eq:Lo} 
\end{equation}
where $O(\cdot, \cdot)$ represents a non-negative counterpart of cosine similarity, promoting orthogonality between the two feature spaces.

Eventually, the four geometric-measure-based regularization terms, addressing both inter- and intra-decoupled spaces, are combined to form the decoupling loss:

\begin{equation} 
L_d = \sum_{k \in \{r,s,m,o\}} \lambda_k L_k, 
\label{eq:Ld} 
\end{equation}
where $\lambda_k$ are weighting coefficients for the individual regularization terms, providing a flexible mechanism to balance their contributions and calibrate the model effectively.

\subsection{Language-Focused Attractor (LFA)}
\label{sec:LFA}
\begin{figure}[t]
  \centering
  \includegraphics[width=1.0\linewidth]{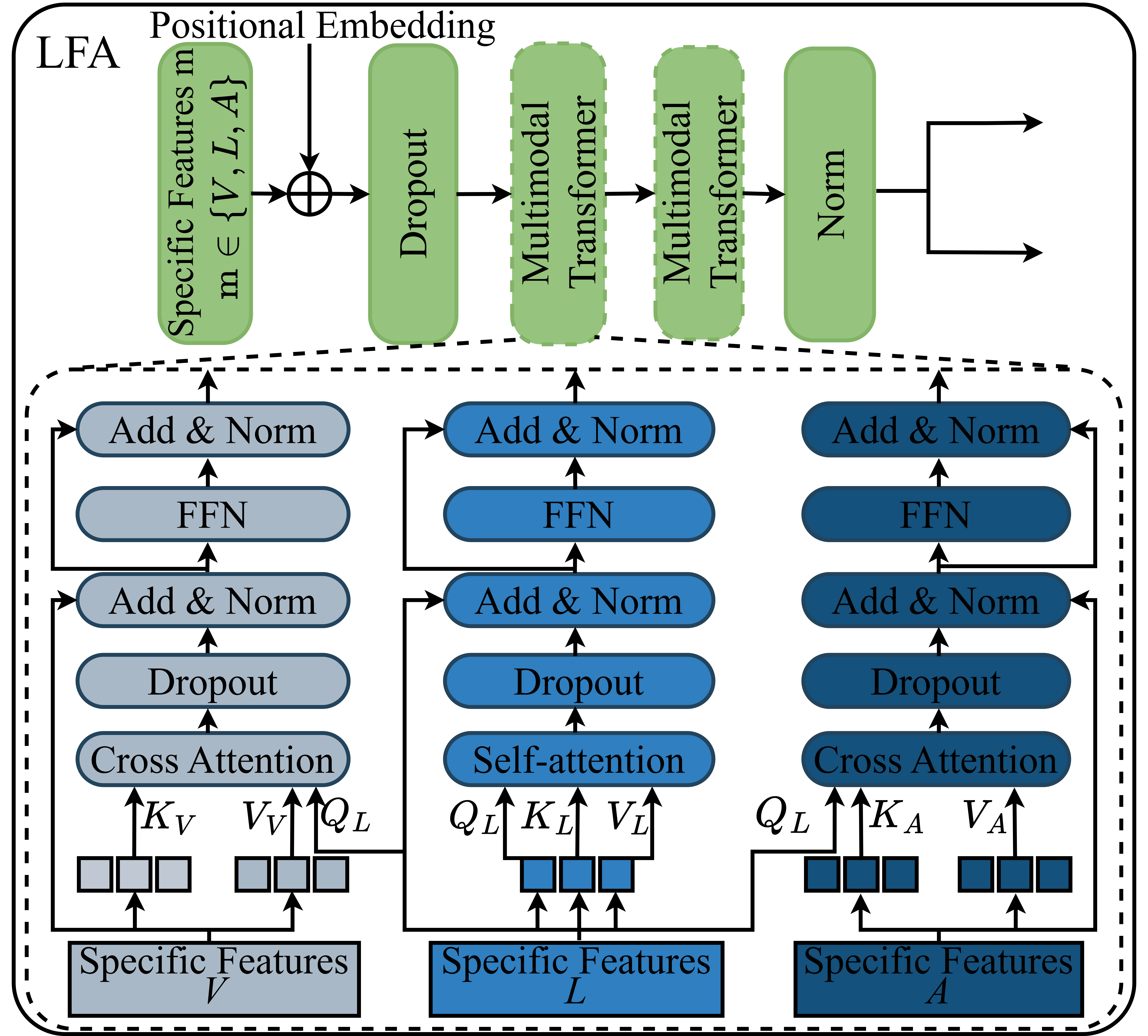}
  \caption{The details of the proposed LFA. The language-focused cross-attention and self-attention achieve targeted feature enhancement: $V$$\rightarrow$$L$, $A$$\rightarrow$$L$, and $L$$\rightarrow$$L$.}
  \label{fig:LFA}
\end{figure}

Unlike conventional feature enhancement methods that aim to bridge modality gaps through cross-modal and graph distillation \cite{gupta2016cross, guo2020online, aslam2023privileged, li2023decoupled}, we propose the LFA in the modality-specific space after feature decoupling. The detailed structure of LFA is depicted in Figure \ref{fig:LFA}.

In the LFA, decoupled modality-specific features $Sp^m$ (where $m \in \{L, V, A\}$, representing language, vision, and audio) are first processed through positional embedding and dropout, then fed into Multimodal Transformer layers. The core operation within these layers is the Multimodal Cross-Attention (MCA) mechanism. LFA performs three branches of MCA, including one self-attention and two cross-attention mechanisms, all centered on the language modality as the \textit{Query} ($Q_L$). This setup allows the language modality to attract complementary information from other modality-specific features $Sp^m$, where $m \in \{L, V, A\}$. The corresponding \textit{Key}-\textit{Value} pairs are defined as ($K_m, V_m$). The MCA operation is mathematically expressed as:

\begin{equation}
\begin{aligned}
  &\text{MCA}(Q_L, K_m, V_m)\\
  =\; &softmax\left( \frac{Q_LK_m^T}{\sqrt{d}}\right)V_m \\
=\; &softmax\left( \frac{(Sp^L)W_{Q_L}(Sp^m)W_{K_m}^T}{\sqrt{d}}\right)(Sp^m)W_{V_m}
\label{eq:MCA}
\end{aligned}
\end{equation}
where $m\in\{L, V, A\}$, $softmax$ represents normalized attention score between $Q_L$ and $K_m$, $W_{Q_L}$ and $W_{K_m}$ are learnable parameters,  $d$ indicates the dimension of $Q_L$ and $K_m$. 
Consequently, the language-focused feature enhancement in the Multimodal Transformer is defined as:
\begin{equation}
\begin{aligned}
  h_m^{n+1} &= LayerNorm(h_m^n + Drop(\text{MCA}(h_m^n)))\\
  h_m^{o} &= LayerNorm(h_m^{n+1} + FFN(h_m^{n+1}))
\end{aligned}
  \label{eq:MT}
\end{equation}
where $m \in \{L, V, A\}$, $Drop(\cdot)$ denotes the Dropout operation, and $FFN(\cdot)$ represents a feed-forward module. Here, $h_m^n$ and $h_m^{n+1}$ are the input and intermediate features, respectively, while $h_m^o$ is the output of a Multimodal Transformer layer. As illustrated in Figure \ref{fig:LFA}, cascaded Multimodal Transformers are employed 
to enhance modality-specific features.

The LFA effectively leverages modality-specific features from three modalities, aligning them with the language modality to strengthen multimodal representation. As shown in Figure \ref{fig:Framework}, the enhanced features are projected into higher-level specific features, denoted as $HSp^m$ ($m \in \{L, V, A\}$), and then integrated with enhanced shared features. Additionally, these features are processed by modality-specific predictors for the specific prediction.

\textbf{Multimodal Fusion.} As illustrated in Figure \ref{fig:Framework}, modality-shared features are first processed through a unified Transformer layer, followed by two fully connected layers, and then projected into higher-level shared features, denoted as $HSh$. Finally, the multimodal fusion layer combines $HSh$ with  $HSp^m$ to form the final multimodal features:

\begin{equation} 
\begin{aligned} 
&F(HSh, HSp^m) \\
=\;& Concat(HSp^L, HSp^V, HSp^A, HSh), 
\end{aligned} 
\end{equation}
where $Concat$ denotes the concatenation operation.

\subsection{Hierarchical Predictions}
\label{sec:predictions}
As shown in Figure \ref{fig:Framework}, a classifier can predict the MSA output $\hat{y}$ after multimodal fusion. The final MSA output loss $L_f$ is defined as:
\begin{equation}
\begin{aligned}
    L_f=\frac{1}{N_d}\sum_{n=0}^{N_d}\lvert\hat{y}_n-y_n\rvert,
\end{aligned}
\end{equation}
where $y_n$ is the MSA label, $N_d$ is the number of samples.
Unlike traditional MSA learning, which only involves a single output loss $L_f$, the proposed DLF explores hierarchical predictions considering modality-shared loss $L_{Sh}$, modality-specific loss $L_{Sp^m}$, and the output loss $L_f$ concurrently. The total MSA learning loss is thus expressed as:
\begin{equation}
\begin{aligned}
    L_{MSA}=\sum_{l \in\{f, Sh, Sp^m\}}\beta_lL_l,
\end{aligned}
\end{equation}
where $m\in\{L, V, A\}$, $\beta_l$ are weighting coefficients that control the relative importance of different losses.

\textbf{Overall Learning Objective.} The proposed DLF framework integrates the decoupling loss $L_d$ and the total MSA learning loss $L_{MSA}$ to form the overall learning objective:

\begin{equation}
\begin{aligned}
    L_{DLF}=L_d + L_{MSA}.
\end{aligned}
\end{equation}

\section{Experiments}
\label{sec:experiments}
\begin{table*}[t]
\centering
\setlength{\tabcolsep}{0.8mm}
\begin{tabular}{lcccccccccccc}
\toprule
\multirow{2}{*}{Method} & \multicolumn{6}{c}{CMU-MOSI} & \multicolumn{6}{c}{CMU-MOSEI} \\
\cmidrule(lr){2-7} \cmidrule(lr){8-13}
 & Acc-7($\uparrow$) & Acc-5($\uparrow$) & Acc-2($\uparrow$) & F1($\uparrow$) & Corr($\uparrow$) & MAE($\downarrow$) & Acc-7($\uparrow$) & Acc-5($\uparrow$) & Acc-2($\uparrow$) & F1($\uparrow$) & Corr($\uparrow$) & MAE($\downarrow$)\\
\midrule
TFN* & 34.90 & 39.39\textsuperscript{\textdagger} & 80.08 & 80.07 & 0.698 & 0.901 & 50.20 & 53.10\textsuperscript{\textdagger} & 82.50 & 82.10 & 0.700 & 0.593 \\
LMF* & 33.20 & 38.13\textsuperscript{\textdagger} & 82.50 & 82.40 & 0.695 & 0.917 & 48.00 & 52.90\textsuperscript{\textdagger} & 82.00 & 82.10 & 0.677 & 0.623 \\
EF-LSTM\textsuperscript{\textdagger} & 35.39 & 40.15 & 78.48 & 78.51 & 0.669 & 0.949 & 50.01 & 51.16 & 80.79 & 80.67 & 0.683 & 0.601 \\
LF-DNN\textsuperscript{\textdagger} & 34.52 & 38.05 & 78.63 & 78.63 & 0.658 & 0.955 & 50.83 & 51.97 & 82.74 & 82.52 & 0.709 & 0.580 \\
MFN\textsuperscript{\textdagger} & 35.83 & 40.47 & 78.87 & 78.90 & 0.670 & 0.927 & 51.34 & 52.76 & 82.85 & 82.85 & 0.718 & 0.575 \\
Graph-MFN\textsuperscript{\textdagger} & 34.64 & 38.63 & 78.35 & 78.35 & 0.649 & 0.956 & 51.37 & 52.69 & 83.48 & 83.43 & 0.713 & 0.575 \\
MulT & 40.00 & 42.68\textsuperscript{\textdagger} & 83.00 & 82.00 & 0.698 & 0.871 & 51.80 & 54.18\textsuperscript{\textdagger} & 82.50 & 82.30 & 0.703 & 0.580 \\
PMR & 40.60 & - & 83.60 & 83.60 & - & - & 52.50 & - & 83.60 & 83.40 & - & - \\
MISA\textsuperscript{\textdagger} & 41.37 & 47.08 & 83.54 & 83.58 & 0.778 & 0.777 & 52.05 & 53.63 & 84.67 & 84.66 & 0.752 & 0.558 \\
MAG-BERT & 43.62 & - & 84.43 & 84.61 & 0.781 & \textbf{0.727} & 52.67 & - & 84.82 & 84.71 & 0.755 & 0.543 \\
DMD** & 46.06 & - & 83.23 & 83.29 & - & 0.752 & 52.78 & - & 84.62 & 84.62 & - & 0.543 \\
\textbf{DLF (Ours)} & \textbf{47.08} & \textbf{52.33} & \textbf{85.06} & \textbf{85.04} & \textbf{0.781} & 0.731 & \textbf{53.90} & \textbf{55.70} & \textbf{85.42} & \textbf{85.27} & \textbf{0.764} & \textbf{0.536} \\
\bottomrule
\end{tabular}
\caption{Comparison on MOSI and MOSEI. Bold is the best. Note: $^{\dagger}$ represents the result from THUIAR's GitHub page \cite{thuiar_mmsa}, $^*$ represents the result from \cite{hazarika2020misa}, - represents the result from the original paper is not provided, and $^{**}$ represents reproduced results from public code with hyper-parameters provided in the original paper.}
\label{tab:comparison}
\end{table*}

\subsection{Datasets and Evaluation Metrics}
We evaluate DLF on two widely used datasets: CMU Multimodal Sentiment Intensity (MOSI) \cite{zadeh2016multimodal} and CMU Multimodal Opinion Sentiment and Emotion Intensity (MOSEI) \cite{zadeh2018multimodal}.

\textbf{MOSI.} The MOSI dataset comprises 2,199 monologue video clips, with audio and visual features extracted at 12.5 Hz and 15 Hz, respectively. The dataset is divided into 1,284 training, 229 validation, and 686 test samples.

\textbf{MOSEI.} The MOSEI dataset, significantly larger, consists of 22,856 movie review video clips sourced from YouTube. Features are extracted at 20 Hz for audio and 15 Hz for visual modalities. The dataset is split into 16,326 training samples, 1,871 validation samples, and 4,659 test samples. For both datasets, each video clip is annotated with a sentiment score ranging from -3 to 3, representing a spectrum from highly negative to highly positive sentiment.

\textbf{Evaluation Metrics.} Consistent with established practices in previous studies \cite{liang2021attention,lv2021progressive,mao2022m}, the performance of MSA is evaluated using multiple metrics: 7-class accuracy (Acc-7), 5-class accuracy (Acc-5), binary accuracy (Acc-2), F1 score, correlation between model predictions and human annotations (Corr), and mean absolute error (MAE). These metrics collectively offer a comprehensive assessment of DLF's effectiveness across various sentiment analysis tasks.

\subsection{Implementation Details}
In this study, we align our methodology with previous works \cite{hazarika2020misa,mao2022m} by utilizing the BERT-base-uncased model \cite{devlin2018bert} to extract unimodal linguistic features. This process generates word representations with a 768-dimensional hidden state. For visual data, DLF employs the Facet framework \cite{baltruvsaitis2016openface} to encode each video frame, focusing on 35 distinct facial action units as detailed in \cite{li2019self}. For audio processing, we utilize the COVAREP framework \cite{degottex2014covarep}, which produces 74-dimensional audio features. Our experiments are implemented using the PyTorch framework and executed on one NVIDIA V100 GPU with 32GB of memory. The model is trained with a batch size of 16 and optimized using an initial learning rate of 1e-4. Early stopping with a patience of 10 epochs is applied to ensure convergence.

\subsection{Main Results}
\textbf{Baselines.} We compare the DLF against eleven leading MSA methods on both benchmarks, including \textbf{EF-LSTM} \cite{williams2018recognizing}, \textbf{LF-DNN} \cite{williams2018dnn}, \textbf{TFN} \cite{zadeh2017tensor}, \textbf{LMF} \cite{liu2018efficient}, \textbf{MFN} \cite{zadeh2018memory}, \textbf{Graph-MFN} \cite{zadeh2018multimodal}, \textbf{MulT} \cite{tsai2019multimodal}, \textbf{PMR} \cite{lv2021progressive}, \textbf{MISA} \cite{hazarika2020misa}, \textbf{MAG-BERT} \cite{rahman2020integrating}, and \textbf{DMD} \cite{li2023decoupled}. 

\textbf{Performance Comparison.} Comparative results, as reported in Table \ref{tab:comparison}, demonstrate that our proposed DLF exhibits superior performance on almost all metrics for both benchmarks. Particularly, we have the following key observations. Compared to decoupled-feature-based MSA methods like MISA \cite{hazarika2020misa}, MulT \cite{tsai2019multimodal}, and DMD \cite{li2023decoupled}, the proposed DLF, especially the LFA, captures effective intermodality dynamics and further improves the multimodal representation capability by enhancing the dominant language in the specific subspace.
Compared to methods that leverage multimodal transformers to learn crossmodal interactions and fusion such as LMF \cite{liu2018efficient}, MFN \cite{zadeh2018memory}, and PMR \cite{lv2021progressive}, our proposed method learns effective multimodal representations in the disentangled subspaces and further facilitates the overall prediction performance using hierarchical predictions which utilizes both pre-fused and post-fused features.

\subsection{Ablation Study}
\label{sec:ab}
We conduct extensive ablation studies to thoroughly examine the impact of various modality combinations, different regularization strategies, and critical components including the Feature Disentanglement Module (FDM), Language-Focused Attractor (LFA), and Hierarchical Predictions (HP).

\textbf{Various Modality Combinations.} As shown in Table \ref{tab:A2}, all analysis are conducted on the MOSI dataset. We first present the performance of each unimodality, it is easy to notice that language modality ($L$) serves as the dominant one.
In the bi-modalities case, we consider both ($L$,$A$) and ($L$,$V$) pairs in our DLF framework, the results not only demonstrate the performance improvement, especially the fine-grained classification, through two modalities but also showcase that language modality attracts useful information from vision ($V$) or audio ($A$) modality by LFA to enhance the multimodal representation capability. Furthermore, the tri-modalities DLF consistently outperforms the bi-modalities DLF on all metrics, which indicates that each modality provides a unique contribution and multimodal learning can effectively improve the MSA performance by reasonably exploiting the information from different modalities.
\begin{table}[t]
\centering
\setlength{\tabcolsep}{1.4mm}
\begin{tabular}{lcccc}
\toprule
Method & Acc-7 (\%) & Acc-2 (\%) & F1 (\%) & MAE($\downarrow$) \\
\midrule
\textbf{DLF (Ours)} & \textbf{47.08} & \textbf{85.06} & \textbf{85.04} & \textbf{0.731} \\
\midrule
\multicolumn{5}{c}{Different Modalities}\\
\midrule
only $A$ & 15.31 & 42.84 & 26.64 & 1.453 \\
only $V$ & 15.01 & 43.29 & 29.73 & 1.455 \\
only $L$ & 45.63 & 84.45 & 84.38 & 0.752 \\
$L$ \& $A$ & 45.77 & 83.84 & 83.88 & 0.741 \\
$L$ \& $V$ & 46.65 & 83.08 & 83.13 & 0.745 \\
\midrule
\multicolumn{5}{c}{Different Regularization}\\
\midrule
w/o $L_r$ & 45.92 & 84.67 & 84.59 & 0.734 \\
w/o $L_s$ & 45.36 & 84.60 & 84.56 & 0.740 \\
w/o $L_m$ & 45.77 & 83.99 & 83.97 & 0.735 \\
w/o $L_o$ & 45.77 & 83.08 & 83.16 & 0.738 \\
\midrule
\multicolumn{5}{c}{Different Components}\\
\midrule
w/o FDM & 45.92 & 84.60 & 84.58 & 0.739 \\
w/o LFA & 42.71 & 83.84 & 83.85 & 0.767 \\
w/o HP & 42.42 & 84.76 & 84.74 & 0.761 \\
\bottomrule
\end{tabular}
\caption{Results of ablation studies on the MOSI benchmark.}
\label{tab:A2}
\end{table}

\textbf{Different Regularization.} We remove each loss to verify the importance of different regularization terms. When removing the soft orthogonality loss $L_o$, the DLF learns decoupled features under the constraints that focus on separate subspaces. The worst performance suggests the importance of the soft orthogonality loss considering the shared and specific subspaces jointly in the feature disentanglement module. Meanwhile, we also notice that the modified triplet loss $L_m$ in the shared subspace improves the overall performance, which indicates the importance of  $L_m$ in learning shared features in the shared subspace. Besides, we observe that both reconstruction loss $L_r$ and specific loss $L_s$ contribute to the model's performance. This is because these two losses ensure feature consistency during disentanglement.

\textbf{Critical Components.} To verify the effectiveness of different components of DLF, we remove each critical component separately. 
The removal of LFA, replaced by three separate \textit{Query} components like MulT \cite{tsai2019multimodal}, leads to a remarkable decrease in overall MSA performance. This demonstrates that the LFA is a straightforward and effective component, mitigating the potential redundancies or conflicts in traditional cross-attention mechanisms.
Moreover, when deactivating the FDM, the performance also becomes inferior to that of DLF, which shows the effectiveness of the designed FDM and further indicates the redundant and conflicting information limits the MSA performance. With a similar phenomenon, subtracting the HP module (which just remains the final output loss function) decreases the overall performance again, revealing the value of both the pre-fused and post-fused features.

\subsection{Further Analysis}
\begin{figure}[t]
    \centering
    \subfigure{
        \includegraphics[width=0.22\textwidth]{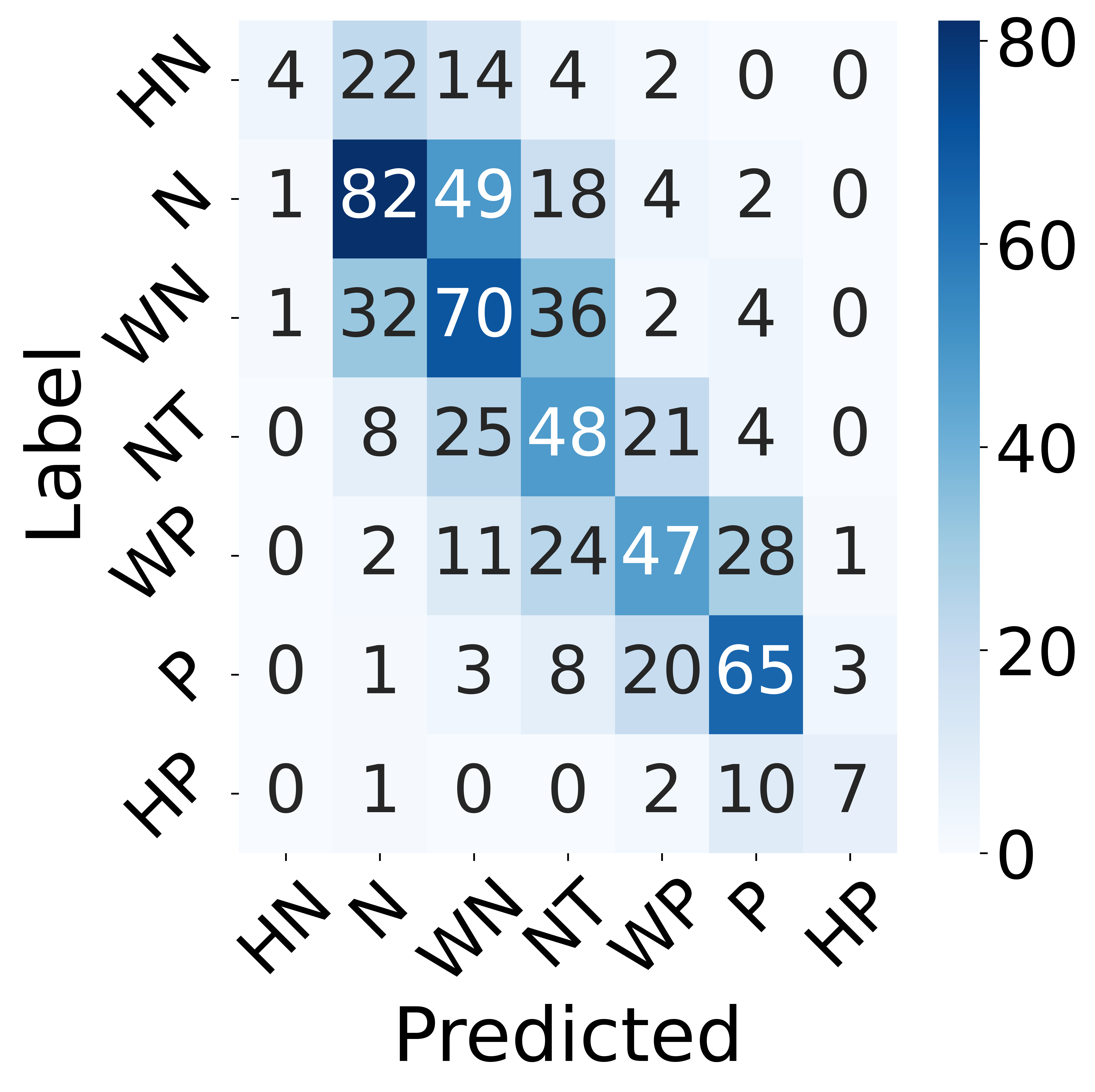}
        \label{fig:first}
    }
    \subfigure{
        \includegraphics[width=0.22\textwidth]{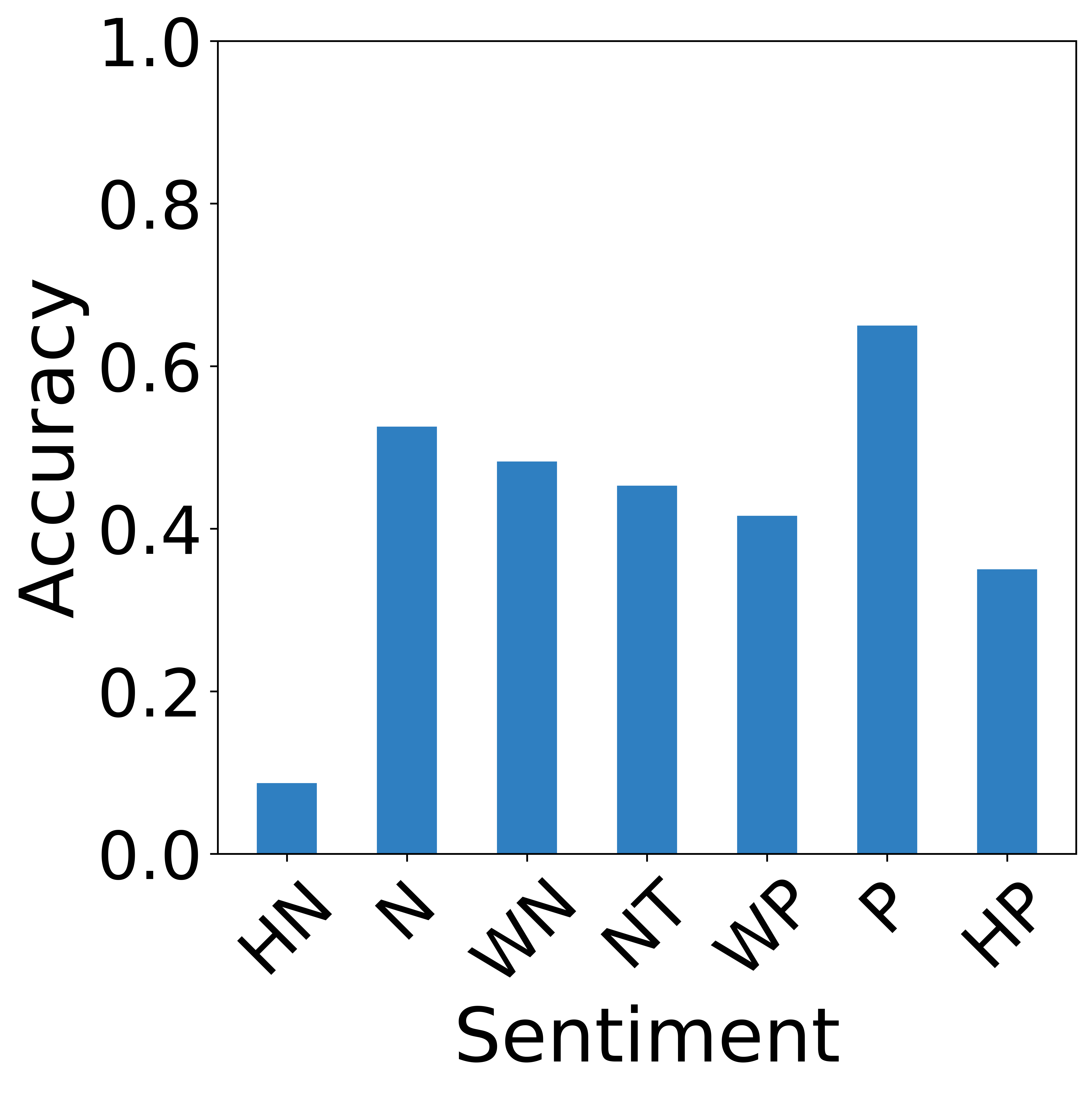}
        \label{fig:second}
    }
    \caption{\textit{Left}: Confusion matrix on MOSI. \textit{Right}: Corresponding accuracy for each sentiment. HN: Highly Negative; N: Negative; WN: Weakly Negative; NT: Neutral; WP: Weak Positive; P: Positive; HP: Highly Positive.}
    \label{fig:matrix}
\end{figure}

To further study the impact of sentiment granularity on MSA, we present the confusion matrix and corresponding accuracy of each sentiment for the MOSI benchmark. As shown in Figure \ref{fig:matrix}, we observe that most sentiment classes have similar accuracy above $40\%$. However, ``HN" and ``HP", especially ``HN", have the worst performance, limiting the overall MSA performance. Furthermore, when we dive into the confusion matrix, it can be noticed that the samples for ``HN" and ``HP" are relatively less than other sentiments, which strongly indicates that the long-tailed distribution of the data limits the overall MSA performance, which can be studied in the future.

\begin{figure}[t]
  \centering
    \includegraphics[width=1.0\linewidth]{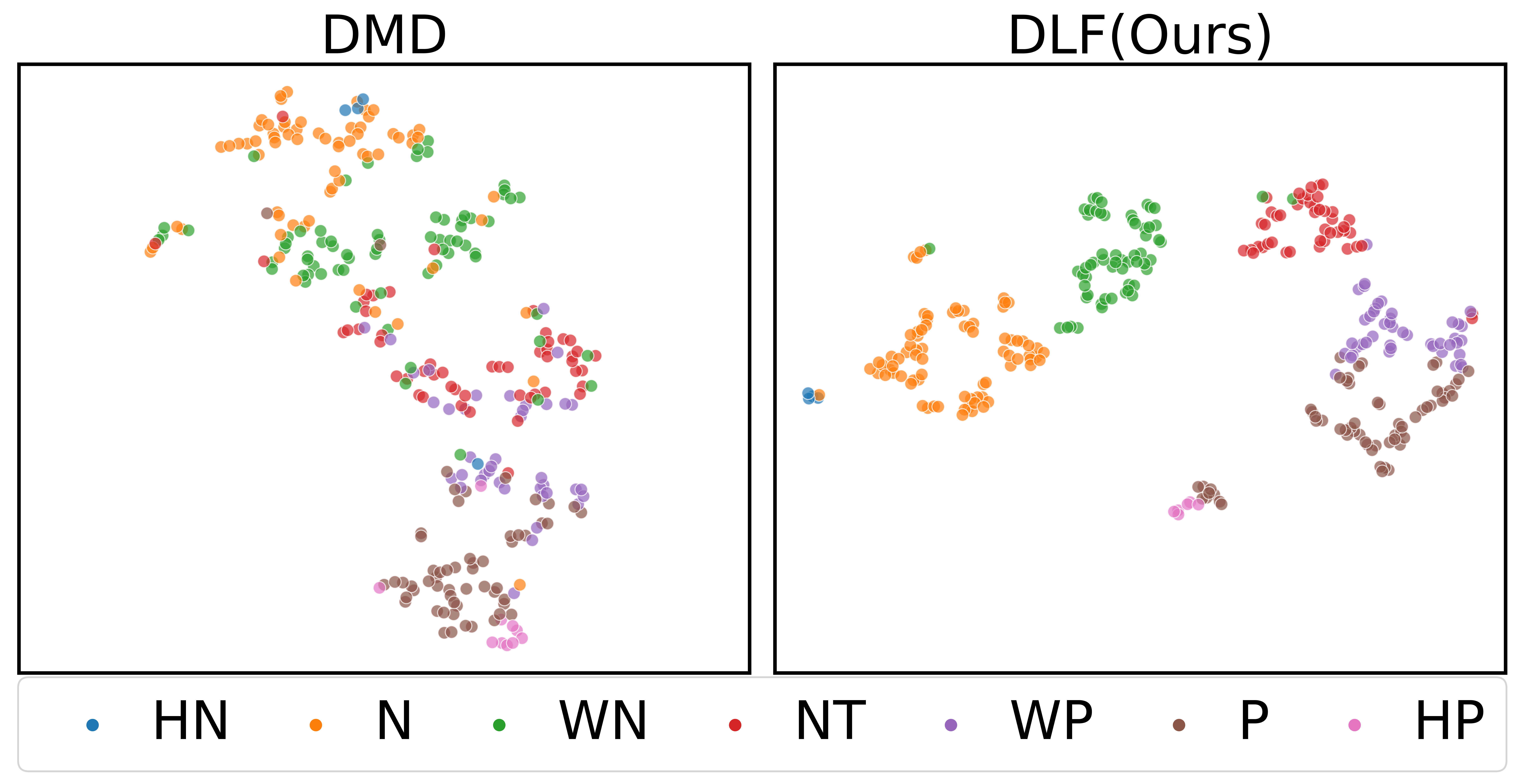}

   \caption{Visualization of the fused multimodal representations. HN: Highly Negative; N: Negative; WN: Weakly Negative; NT: Neutral; WP: Weak Positive; P: Positive; HP: Highly Positive.}
   \label{fig: tsne}
\end{figure}

To better understand the effectiveness of our method, we visualize the distribution of fused multimodal representations. As depicted in Figure \ref{fig: tsne}, compared to DMD, which is a decoupled-multimodal-distillation strategy, our proposed DLF shows superior performance in separating different sentiments. This is mainly due to that the LFA mitigates redundant and conflicting information during multimodal interaction compared to the interaction between random pairs.

\section{Conclusion}
In this paper, we propose the DLF framework to improve the MSA performance. DLF yields powerful multimodal representations by following the pipeline of feature extraction, disentanglement, enhancement, fusion, and prediction, mainly benefiting from the feature disentanglement module, language-focused attractor, and hierarchical predictions. Extensive results verify the superiority of DLF by comparisons with eleven baselines and comprehensive ablation studies.

\textbf{Broad impacts.} (i) This study demonstrates its potential to exploit the imbalanced capabilities of various modalities in multimodal learning, thereby setting a new benchmark in this field. (ii) The proposed LFA facilitates the generalization of our method to other multimodal scenarios by changing the dominant modality.
\textbf{Limitation and future work.} Our method only considers the scenarios of complete modalities. When facing missing modalities, the feature disentanglement and enhancement modules are potentially limited.

\section{Acknowledgments}
This work is supported in part by NIH R01EB033387, NSF CNS-2122320, and NSF CCF-2324937. Tianlong Chen is supported by NIH OT2OD038045-01 and UNC SDSS Seed Grant.

\bibliography{aaai25}
\end{document}